# QuPCG: Quantum Convolutional Neural Network for Detecting Abnormal Patterns in PCG Signals

Yasaman Torabi [1,*], Shahram Shirani [1,2], and James P. Reilly [1]

[1] Department of Electrical and Computer Engineering, McMaster University, Hamilton, L8S 4K1, Canada;
[2] L.R. Wilson/Bell Canada Chair in Data Communications, Hamilton, L8S 4L7, Canada
* Correspondence: torabiy@mcmaster.ca

**Abstract.** Early identification of abnormal physiological patterns is essential for the timely detection of cardiac disease. This work introduces a hybrid quantum–classical convolutional neural network (QCNN) designed to classify S3 and murmur abnormalities in heart sound signals. The approach transforms one-dimensional phonocardiogram (PCG) signals into compact two-dimensional images through a combination of wavelet feature extraction and adaptive threshold compression methods. We compress the cardiac sound patterns into an 8-pixel image so that only 8 qubits are needed for the quantum stage. Preliminary results on the HLS-CMDS dataset demonstrate 93.33% classification accuracy on the test set, and 97.14% on the train set, suggesting that quantum models can efficiently capture temporal–spectral correlations in biomedical signals. To our knowledge, this is the first application of a QCNN algorithm for bioacoustic signal processing. The proposed method represents an early step toward quantum-enhanced diagnostic systems for resource-constrained healthcare environments.

## 1 Introduction

Cardiovascular diseases remain the major cause of death across the world, with more than 19 million fatalities each year [1]. Early recognition of cardiac disorders is essential for preventing severe complications and reducing mortality [2]. Standard diagnostic methods such as electrocardiography, echocardiography, and auscultation are reliable but depend on clinical experience and advanced instruments [3]. These factors limit their use in rural or low-resource regions. Automatic analysis of heart sounds has therefore become an active area of research for affordable screening [4]. Heart sounds are recorded as phonocardiograms (PCGs). They contain temporal and spectral patterns linked to valve activity and cardiac rhythm. Classical signal-processing techniques, such as Fourier and wavelet transforms, have been used to extract features from PCGs [5]. Deep learning methods have improved this field by improving accuracy under noise or overlapping conditions. Convolutional neural networks (CNNs) and recurrent neural networks (RNNs) can detect murmurs with high precision [6], [7]. CNN models convert

PCG signals into spectrograms or wavelet scalograms and learn spatial patterns that correspond to cardiac pathologies. Transfer learning and attention mechanisms have further improved recognition rates on small datasets [8]. However, such networks still require strong computational hardware. These requirements restrict their real-time use on portable diagnostic devices [9]. Quantum computing provides a new way to process data through qubits that can represent several states simultaneously. Quantum machine learning (QML) combines this property with classical optimization to improve computational efficiency [10]. In hybrid networks, quantum circuits extract correlations in feature space while classical layers update the parameters. Variational quantum circuits (VQCs) and hybrid quantum neural networks have been applied to pattern classification, clustering, and feature selection [11], [12]. Quantum convolutional neural networks (QCNNs) extend this idea by using layers of quantum convolution and pooling gates that mimic hierarchical feature extraction [13]. These models have been used for biomedical classifications, such as breast cancer diagnosis [14], and electrocardiogram (ECG) signal recognition [15]. Despite this progress, no earlier study has applied QCNNs to heart sounds. PCG signals differ from images and ECG traces because they contain short acoustic pulses with complex frequency variation. A quantum model must therefore compress and encode the information efficiently before processing. This study aims to demonstrate that a compact hybrid model can detect cardiac abnormalities with limited computational cost. Fig. 1 illustrates our proposed method.

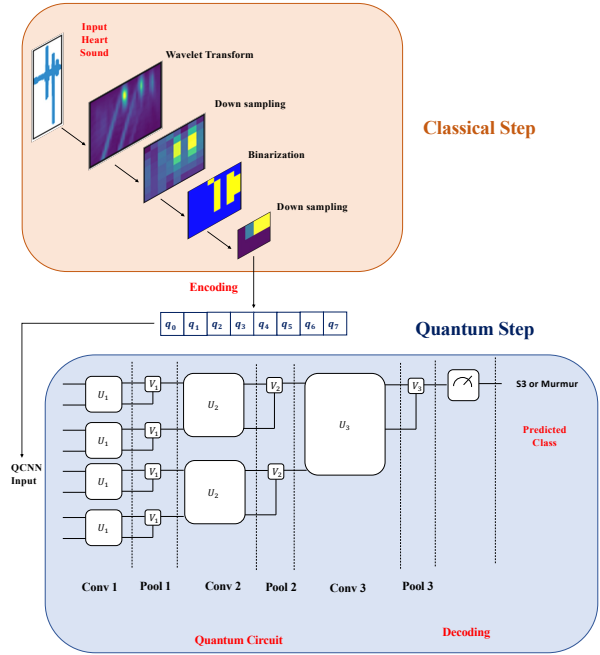

**Fig. 1** Overview of the proposed hybrid quantum–classical model for heart sound classification. The classical step applies wavelet transform, downsampling, and binarization to generate compact quantum-ready 8-qubit feature maps. The quantum step encodes these features into qubits and processes them through three stages of successive quantum convolution and pooling layers, with the final measurement used to predict the class (S3 or murmur).

The main contributions are:

- **Transforming sound signals into images, which enables image processing methods for pattern recognition.**

- **Introducing a compression pipeline using Wavelet transform, downsampling, and binarization that compresses images into eight pixels suitable for 8-qubit quantum encoding.**

- **Designing a quantum convolutional neural network that learns hierarchical features on Qiskit simulators.**

## 2 Related Work

Early cardiac disease studies used conventional digital signal processing methods such as the Fourier transform to extract time–frequency information from phonocardiograms (PCGs) [16]. These techniques produced acceptable accuracy on controlled recordings but often failed on noisy or overlapping signals. To overcome these limitations, researchers applied traditional machine learning models for PCG classification. Support vector machines (SVMs), random forests, non-negative matrix factorization (NMFs), and k-nearest neighbors (KNN) were trained using features such as spectral entropy, zero-crossing rate, and Mel-frequency coefficients [17]. These models achieved moderate success but required manual feature selection and tuning. The introduction of deep learning improved the performance of heart sound recognition systems. Springer et al. [18] proposed an automated segmentation method using hidden semi-Markov models, which became a reference for separating S1 and S2 cardiac components. Gosh et al. [19] implemented a convolutional neural network (CNN) that classified murmurs from time–frequency images. Vinay et al. [20] provided a recurrent neural network-based bidirectional long short-term memory with a generative adversarial networks approach (RNN–BiLSTM–GAN) to improve feature optimization and classification accuracy in PCG-based cardiovascular disease detection. Singh et al. [21] proposed an ensemble-based transfer learning model trained on spectrogram images from PCG signals to address data imbalance. These works established deep learning as a reliable tool for heart sound recognition. Although deep learning methods have achieved good results, their performance depends on the size and diversity of labeled datasets. Large neural networks require extensive data, computation, and energy, which limits their use in portable medical devices [22]. Several studies have explored lightweight or compressed architectures to address this issue. For example, Huang et al. [23] developed an adaptive temporal compression technique that reduces computational complexity while preserving essential dynamics.

Quantum computing offers a new way to reduce model complexity and data requirements. Quantum machine learning (QML) combines the probabilistic properties of quantum systems with the optimization capability of classical algorithms [24]. In QML models, information is encoded into qubits that can exist in multiple states at once. Variational quantum circuits (VQCs) are one of the most studied architectures, where circuit parameters are trained using classical optimizers. These models can represent high-dimensional data with fewer trainable parameters [25]. Hybrid quantum–classical neural networks (HQNNs) extend this concept by connecting quantum layers with classical preprocessing or postprocessing blocks. Cerezo et al. [26] demonstrated that hybrid variational algorithms can solve pattern recognition problems efficiently when data are encoded in a suitable basis. Quantum convolutional neural networks (QCNNs) generalize the concept of convolution to quantum information processing. Ullah et al. [27] proposed a fully connected quantum convolutional neural network (FCQ-CNN) for ischemic heart disease classification, demonstrating that quantum circuit–based architectures can achieve higher accuracy and reduced parameter complexity compared to classical CNN models. Cong et al. [28] introduced the QCNN architecture inspired by the multi-scale entanglement renormalization ansatz, which performs local unitary operations followed by pooling layers that reduce the number of qubits while preserving correlations. The QCNN design allows hierarchical feature extraction similar to that of classical CNNs but with fewer resources. Recent research has shown that QCNNs can be used in several scientific domains. Zhang et al. [29] used a QCNN for electrocardiogram (ECG) signal classification. Li et al. [30] introduced a quantum-inspired scalable convolutional neural network for pneumonia diagnosis that integrates parallel quantum feature extractors on medical imaging datasets. Similar studies applied QCNNs to molecular property prediction [31] and quantum chemistry [32]. To our best knowledge, no earlier work has applied QCNNs to heart sound signals. PCGs differ from ECG and image data because they include transient acoustic components with strong frequency modulation. Classical CNNs often require large filter sizes to capture these features, which increases memory usage and computation. A QCNN can extract multi-level temporal–spectral correlations in a smaller feature space through quantum entanglement. Some related work has focused on hybrid quantum models for other biosignals, such as electroencephalogram (EEG) classification [33], and stress detection [34]. These results indicate that quantum models can handle physiological data effectively when designed with proper encoding and compression. The current study extends this direction by presenting a QCNN for heart sound analysis. The method transforms PCG segments into wavelet representations. The data are compressed to eight pixels, which are encoded into qubits. Quantum convolution and pooling layers extract correlations among the qubits, and a classical optimizer updates the circuit parameters. This design offers a step toward quantum-assisted biomedical diagnostics.

## 3      Theoretical Background

Heart-sound recordings are non-stationary signals whose frequency content changes with time. The wavelet transform provides time–frequency information for signal representation, which expresses a signal $s(t)$ as scaled and shifted versions of a basic waveform, called the mother wavelet (Eq. 1).

$$W(a,b) = \int s(t)\, \psi^* \left(\frac{t-b}{a}\right) dt, \qquad (1)$$

where $a$ is the scale, $b$ is the translation parameter, and $\psi(t)$ is the mother wavelet. This produces a two-dimensional representation of the heart-sound signals suitable for image-based analysis. Convolutional neural networks (CNNs) extract patterns from these maps through local filtering and pooling. Each convolution layer computes the correlation between an input map $x$ and a kernel $w$. Quantum machine learning (QML) extends these operations to quantum space. The information is encoded into qubits that exist in superposition (Eq. 2). The information is then processed through parameterized unitary gates $U(\theta)$ that evolve the state (Eq. 3). Measurement of an observable $O$ gives an expected value (Eq. 4), which serves as the network output. In hybrid quantum–classical models, these expectation values feed classical optimizers that adjust the rotation vector $\theta$ to minimize loss. This mechanism allows compact circuits to model complex nonlinear relationships beyond classical feature spaces.

$$|\psi\rangle = \alpha|0\rangle + \beta|1\rangle, \qquad |\alpha|^2 + |\beta|^2 = 1. \tag{2}$$

$$|\psi'\rangle = U(\theta)|\psi\rangle. \tag{3}$$

$$\langle O \rangle = \langle \psi'|O|\psi'\rangle. \tag{4}$$

## 4 Methodology

### 4.1 Dataset

We used the heart sounds dataset (HLS-CMDS) collected from clinical manikins using the 3M Littmann CORE Digital Stethoscope. The dataset contains 535 recordings from a CAE Juno manikin, including both normal and abnormal cardiac sounds. Recordings were captured at 22,050 Hz for 15-second segments in a quiet simulation environment to reduce background noise. The manikin sounds originate from real patient recordings. We placed the stethoscope on standard auscultation landmarks and kept it steady to minimize noise. Our dataset is publicly available, with details of the recording device, sensor placement, environment, and annotations provided in [35].

### 4.2 Segmentation

In a phonocardiogram (PCG), the first and second heart sounds (S1, S2) correspond to the closure of the heart valves, marking the start and end of each cardiac cycle. The third heart sound (S3) appears as a low-frequency vibration after S2, indicating heart failure in most cases, while murmurs present as prolonged oscillations, often associated with blood backflow [36]. We focused on the classification of S3 versus murmur abnormal sounds. We segmented the recordings based on cardiac cycles (Fig. 2). Each signal was resampled to 4 kHz and analyzed to identify the heart-sound peaks. The distance between peaks was adjusted in order to match the average duration of a single cardiac cycle. Segmentation boundaries were defined at the midpoints between consecutive peaks, ensuring that each extracted segment captured one complete cardiac cycle.

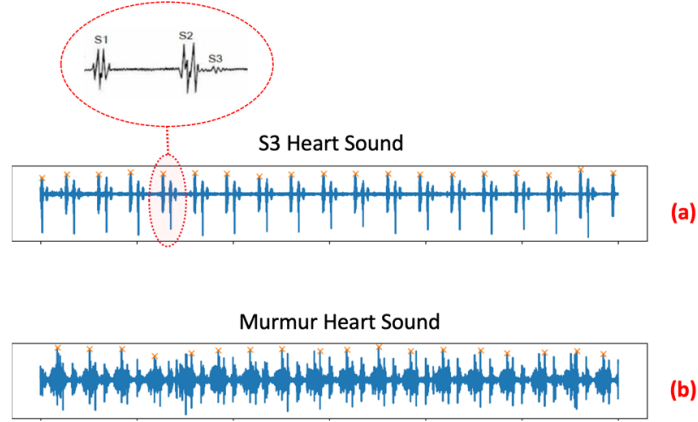

**Fig. 2** PCG signal segments: **(a)** heart sound with additional S3, and **(b)** murmur heart sound.

### 4.3 From Time-Series Signals to Energy Map Images

We transformed each segment into a two-dimensional time–frequency representation using the continuous wavelet transform (CWT) with a complex morlet (cmor) mother wavelet. We applied 128 scales to compute the scalogram and visualized the magnitude of the complex coefficients as an image. Unlike conventional approaches that process long time-series data, our method converts heart sounds into compact wavelet images, allowing us to apply image-based pattern recognition techniques directly. This transformation enables effective compression, which is an essential advantage when operating with a limited number of qubits. The wavelet-based representation also preserves transient events such as murmur patterns more effectively. The scalograms were resized to $32 \times 32$ pixels to provide a consistent input format for the next compression stage.

### 4.4 Feature Compression

As shown in Fig. 3, we compress each $32 \times 32$ scalogram into a compact quantum-ready format. Max-pooling with $4 \times 4$ kernel downsamples the image to $8 \times 8$. We binarize the map into high and low-energy regions so that the patterns remain visible while weak fluctuations are suppressed. The $8 \times 8$ maps are reduced to eight representative values to align with the 8-qubit architecture.

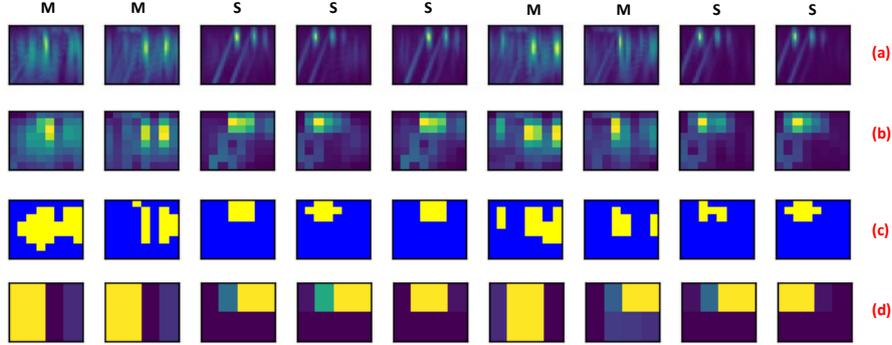

**Fig. 3** Progressive compression of wavelet scalograms for quantum encoding: **(a)** original 32 × 32 time–frequency images of murmur (M) and S3 (S) sounds, **(b)** 8 × 8 max-pooled maps, **(c)** binarized energy patterns highlighting dominant regions, and **(d)** final 8-value representations matched to the 8-qubit QCNN input format.

### 4.5 Quantum Encoding and QCNN Design

We mapped each normalized pixel intensity to the rotation angle of a single-qubit gate. Eight pixel values were encoded into eight qubits. Fig. 4a shows the feature-mapping circuit implemented, where each qubit undergoes Hadamard and phase-rotation gates to embed classical image features into the quantum state space. The QCNN contained three sets of alternating convolutional and pooling layers. Fig. 4b illustrates parametrized unitary gates $U(\theta)$ used in the convolution circuit. We apply a pooling layer after the convolutional layer to reduce the dimensions of the quantum circuit. Fig. 4c shows the two-qubit pooling circuit $V(\theta)$. This layer merges the information of two qubits into one by first applying a unitary operation that transfers and encodes data from one qubit to the other, after which the second qubit is discarded and excluded from further processing or measurement. We apply this two-qubit circuit to different pairs of qubits to create a pooling layer for 8 qubits. The final measured qubit provided the classification feature. After each quantum execution, the expectation value of the Pauli-Z observable was measured, and the loss between predicted and actual labels was computed to update the gate parameters iteratively until convergence.

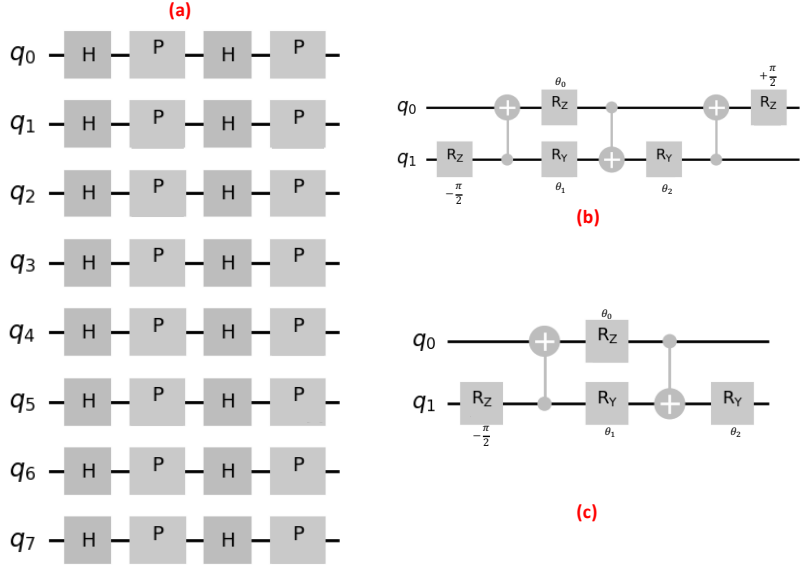

**Fig. 4** Quantum convolutional neural network components: **(a)** Feature-mapping circuit, **(b)** parametrized two-qubit unitary circuit used in the convolutional layer, and **(c)** parametrized two-qubit unitary circuit used in the pooling layer.

## 4.6 Settings

We trained the model in a hybrid loop that combined quantum simulation and classical optimization. The circuit was implemented in Qiskit version 0.45 using the AerSimulator backend. The COBYLA optimizer was used with a learning rate of 0.01, batch size of 16, and 200 epochs. We executed the experiments on a GPU server equipped with an Nvidia GeForce RTX 4090 16384, 24 GB. The network employed 8 qubits with a circuit depth of 3 layers and used the COBYLA optimizer with a learning rate of 0.01. This configuration provided stable convergence and efficient simulation within limited quantum resources.

## 4.7 Experimental Results

We experimented with three approaches for converting PCG time-series signals into two-dimensional representations suitable for the QCNN. Table 1 compares the performance of different QCNNs under various signal-to-image preprocessing strategies. In the first method (**I-QuPCG**), the raw time-series signals were directly converted into grayscale images; however, this approach yielded poor performance, as the temporal information was largely lost during compression. The second approach (**M-QuPCG**) used Short-Time Fourier Transform (STFT) to generate Mel spectrograms, which captured more meaningful frequency patterns and slightly improved accuracy. Finally, the **W-QuPCG** applied a wavelet transform to obtain time–frequency maps that preserved the features of biomedical signals more effectively. In addition, W-QuPCG demonstrates the fastest convergence and achieves the highest accuracy with the lowest loss, confirming its superior learning efficiency, as shown in Fig. 5. Building upon this, we further increased the number of training epochs to 1000, resulting in the enhanced **W-QuPCG[+]**, which achieved the best overall performance and demonstrated the optimal configuration of the proposed model.

**Table 1.** Performance comparison of different implemented QCNNs (mean ± std).

| Method | Accuracy (Train) | Accuracy (Test) | Loss (Train) | Loss (Test) |
| --- | --- | --- | --- | --- |
| I-QuPCG | 51.06 ± 2.3 % | 47.62 ± 2.8 % | 1.00 ± 0.25 | 1.10 ± 0.54 |
| M-QuPCG | 74.29 ± 5.2 % | 53.33 ± 4.0 % | 0.80 ± 0.13 | 0.91 ± 0.25 |
| W-QuPCG | 91.43 ± 2.9 % | 80.00 ± 6.1 % | 0.69 ± 0.12 | 0.73 ± 0.03 |
| **W-QuPCG[+]** | 97.14 ± 4.6 % | 93.33 ± 2.9 % | 0.42 ± 0.62 | 0.45 ± 0.12 |

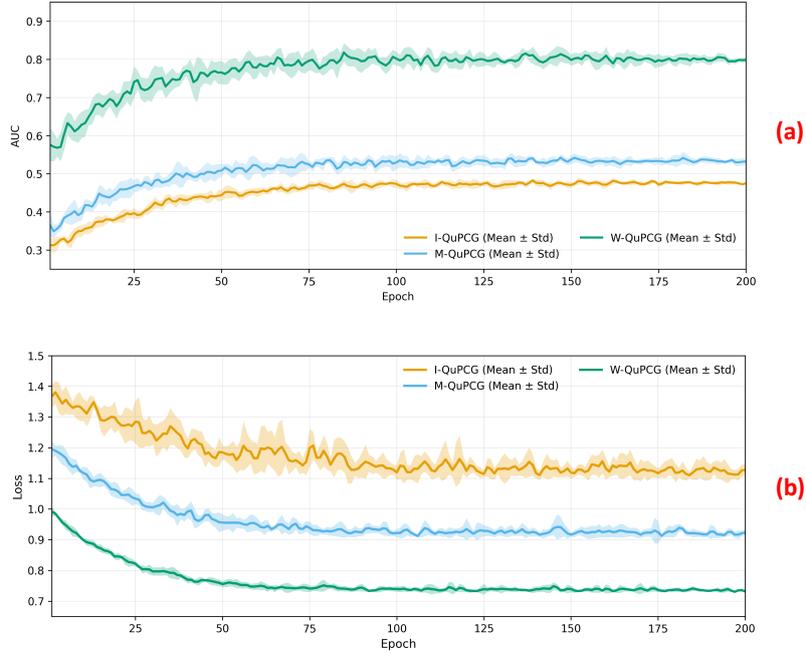

**Fig. 5** Performance analysis of different QCNNs: **(a)** accuracy, and **(b)** loss function value.

## 5   Discussion

The results indicate that the hybrid quantum–classical model can distinguish abnormal heart sound patterns using very limited pixels. The wavelet transformation converts signals into energy maps, which allow the quantum circuit to analyze correlations that are difficult to capture in the time domain. The 8-qubit QCNN performed an acceptable separation between murmur and normal sounds. This supports the idea that small-scale quantum models can perform pattern recognition tasks efficiently when guided by well-engineered preprocessing and compression stages. Although the study demonstrates promising performance, there are practical challenges that must be addressed. The use of simulated backends cannot capture all physical noise sources that would occur on real quantum hardware. Future work should extend this research to real devices and compare the effect of decoherence on model stability. The dataset size is another limitation since quantum models benefit from diverse and balanced samples. Expanding the dataset to include multiple cardiac conditions and patient populations will improve generalization. In future studies, additional encoders, adaptive pooling circuits, and integration with explainable modules may help to clarify how quantum layers learn medical features.

## 6      Conclusion

This work presented a hybrid quantum–classical convolutional network for the classification of abnormal heart sound patterns. The method combines wavelet-based feature extraction with adaptive compression and quantum encoding to create a compact learning model. The results show that even with a small number of qubits, quantum circuits can process biomedical data effectively when designed with domain-specific constraints. These findings support the potential of quantum learning for diagnostic applications and motivate further investigation using real quantum processors and larger clinical datasets.

## 7      Dataset and Source Code

The python scripts are publicly available on https://github.com/Torabiy/QuPCG. The HLS-CMDS dataset is publicly available on https://github.com/Torabiy/HLS-CMDS, with details of the recording device, sensor placement, environment, and annotations provided in [35].

## References

[1] World Health Organization: Cardiovascular diseases (CVDs). Fact Sheet, July 31 2025. https://www.who.int/news-room/fact-sheets/detail/cardiovascular-diseases-(cvds)(accessed on 15 October 2025).

[2] Almansouri, N.E., Awe, M., Rajavelu, S., Jahnavi, K., Shastry, R., Hasan, A., Hasan, H., Lakkimsetti, M., AlAbbasi, R.K., Gutiérrez, B.C., Haider, A.: Early Diagnosis of Cardiovascular Diseases in the Era of Artificial Intelligence: An In-Depth Review. Cureus 16(3), e55869 (2024). https://doi.org/10.7759/cureus.55869

[3] Gudigar, A., et al.: *Automated System for the Detection of Heart Anomalies Using Phonocardiograms: A Systematic Review.* IEEE Access 12, 138399–138428 (2024). https://doi.org/10.1109/ACCESS.2024.3465511

[4] Krones, F., Walker, B.: *From theoretical models to practical deployment: A perspective and case study of opportunities and challenges in AI-driven cardiac auscultation*

*research for low-income settings.* PLOS Digital Health 3(12), e0000437 (2024). https://doi.org/10.1371/journal.pdig.0000437

[5] Kannan, A., Saikia, M.J., Kumar, S., Datta, S.: *Detection of Valvular Heart Diseases From PCG Signals Using Machine and Deep Learning Models: A Review.* IEEE Access 13, 110344–110364 (2025). https://doi.org/10.1109/ACCESS.2025.3583263

[6] Patwa, A., Rahman, M. M. U., Al-Naffouri, T. Y. *Heart Murmur and Abnormal PCG Detection via Wavelet Scattering Transform and 1D-CNN.* IEEE Sensors Journal, 25(7), 12430–12443 (2025). https://doi.org/10.1109/JSEN.2025.3541320

[7] Hsieh, Y.-T., Lin, P.-C., Chen, H.-W., et al. *Development and Validation of an Integrated Residual-Recurrent Neural Network Model for Automated Heart Murmur Detection in Pediatric Populations.* Scientific Reports 15(1), 19155 (2025). https://doi.org/10.1038/s41598-025-19155-8

[8] Alotaibi, A., and AlSaeed, D. Skin cancer detection using transfer learning and deep attention mechanisms. Diagnostics 15(1), 99 (2025). https://doi.org/10.3390/diagnostics15010099

[9] Gholizade, M., Soltanizadeh, H., Rahmanimanesh, M., Sana, S. A review of recent advances and strategies in transfer learning. International Journal of System Assurance Engineering and Management, 1–40 (2025). https://doi.org/10.1007/s13198-025-02220-1

[10] Patil, R.Y., Patil, Y.H., Doss, S. Utilizing Quantum Computing. In: *The Rise of Quantum Computing in Industry 6.0 Towards Sustainability: Revolutionizing Smart Disaster Management*, pp. 141 (2024). https://doi.org/10.1007/978-981-97-9924-8_9

[11] Chen, Y. A novel image classification framework based on variational quantum algorithms. Quantum Information Processing 23(10), 362 (2024). https://doi.org/10.1007/s11128-024-04487-4

[12] Wang, A., Hu, J., Zhang, S., Li, L. Shallow hybrid quantum-classical convolutional neural network model for image classification. Quantum Information Processing 23(1), 17 (2024). https://doi.org/10.1007/s11128-023-04097-9

[13] Long, C., Huang, M., Ye, X., Futamura, Y., Sakurai, T. Hybrid quantum-classical-quantum convolutional neural networks. Scientific Reports 15(1), 31780 (2025). https://doi.org/10.1038/s41598-025-31780-2

[14] Zhang, S., Wang, A., Li, L. Quantum-convolution-based hybrid neural network model for arrhythmia detection. Quantum Machine Intelligence 6(2), 75 (2024). https://doi.org/10.1007/s42484-024-00142-7

[15] Xiang, Q., Li, D., Hu, Z., Yuan, Y., Sun, Y., Zhu, Y., Fu, Y., Jiang, Y., Hua, X. Quantum classical hybrid convolutional neural networks for breast cancer diagnosis. Scientific Reports 14(1), 24699 (2024). https://doi.org/10.1038/s41598-024-24699-4

[16] S. K. Ghosh, R. K. Tripathy and P. R. N, "Classification of PCG Signals using Fourier-based Synchrosqueezing Transform and Support Vector Machine," 2021 IEEE Sensors, Sydney, Australia, 2021, pp. 1–4, doi: 10.1109/SENSORS47087.2021.9639687.

[17] Torabi, Y., Shirani, S., & Reilly, J. P. (2025). Large language model-based nonnegative matrix factorization for cardiorespiratory sound separation. arXiv preprint arXiv:2502.05757. https://doi.org/10.48550/arXiv.2502.05757

[18] D. B. Springer, L. Tarassenko and G. D. Clifford, "Logistic Regression-HSMM-Based Heart Sound Segmentation," IEEE Transactions on Biomedical Engineering, vol. 63, no. 4, pp. 822–832, Apr. 2016, doi: 10.1109/TBME.2015.2475278.

[19] S. K. Ghosh, R. N. Ponnalagu, R. K. Tripathy, G. Panda and R. B. Pachori, "Automated Heart Sound Activity Detection From PCG Signal Using Time–Frequency-Domain Deep Neural Network," IEEE Transactions on Instrumentation and Measurement, vol. 71, pp. 1–10, 2022, Art no. 4006710, doi: 10.1109/TIM.2022.3192257.

[20] N. A. Vinay, K. N. Vidyasagar, S. Rohith, D. Pruthviraja, S. Supreeth and S. H. Bharathi, "An RNN-Bi LSTM Based Multi Decision GAN Approach for the Recognition of Cardiovascular Disease (CVD) From Heart Beat Sound: A Feature Optimization Process," *IEEE Access*, vol. 12, pp. 65482–65502, 2024, doi: 10.1109/ACCESS.2024.3397574.

[21] Singh, S.A., Devi, N.D., Singh, K.N. et al. An ensemble-based transfer learning model for predicting the imbalance heart sound signal using spectrogram images. Multimed Tools Appl 83, 39923–39942 (2024). https://doi.org/10.1007/s11042-023-17186-9

[22] MohiEldeen Alabbasy, F., Abohamama, A., & Alrahmawy, M. F. (2023). Compressing medical deep neural network models for edge devices using knowledge distillation. *Journal of King Saud University - Computer and Information Sciences*, *35*(7), 101616. https://doi.org/10.1016/j.jksuci.2023.101616

[23] Huang, H., Wang, Y., Cai, M., Wang, R., Wen, F., & Hu, X. (2024). Adaptive temporal compression for reduction of computational complexity in human behavior recognition. Scientific Reports, 14(1), 1-11. https://doi.org/10.1038/s41598-024-61286-x

[24] Zeguendry, A., Jarir, Z., & Quafafou, M. (2023). Quantum Machine Learning: A Review and Case Studies. *Entropy*, *25*(2), 287. https://doi.org/10.3390/e25020287

[25] Yetiş, H., & Karaköse, M. (2023). Variational quantum circuits for convolution and window-based image processing applications. Quantum Science and Technology, 8(4), 045004.

[26] Cerezo, M., Arrasmith, A., Babbush, R., Benjamin, S. C., Endo, S., Fujii, K., ... & Coles, P. J. (2021). Variational quantum algorithms. Nature Reviews Physics, 3(9), 625-644.

[27] U. Ullah, A. G. O. Jurado, I. D. Gonzalez and B. Garcia-Zapirain, "A Fully Connected Quantum Convolutional Neural Network for Classifying Ischemic Cardiopathy," in IEEE Access, vol. 10, pp. 134592-134605, 2022, doi: 10.1109/ACCESS.2022.3232307.

[28] Cong, I., Choi, S., & Lukin, M. D. (2019). Quantum convolutional neural networks. *Nature Physics*, *15*(12), 1273-1278. https://doi.org/10.1038/s41567-019-0648-8

[29] Zhang, S., Wang, A., & Li, L. (2024). Quantum-convolution-based hybrid neural network model for arrhythmia detection. Quantum Machine Intelligence, 6(2), 75.

[30] Li, D., Sun, Y., Yuan, Y., Hu, Z., Xiang, Q., Jiang, Y., ... & Hua, X. (2026). A quantum-inspired medical scalable convolutional neural network for Intelligent pneumonia diagnosis. Biomedical Signal Processing and Control, 112, 108440.

[31] Lu, C., Liu, Q., Wang, C., Huang, Z., Lin, P., & He, L. (2019, July). Molecular property prediction: A multilevel quantum interactions modeling perspective. In Proceedings of the AAAI conference on artificial intelligence (Vol. 33, No. 01, pp. 1052-1060).

[32] Motlagh, D., Lang, R. A., Jain, P., Campos-Gonzalez-Angulo, J. A., Maxwell, W., Zeng, T., ... & Arrazola, J. M. (2025). Quantum algorithm for vibronic dynamics: case study on singlet fission solar cell design. Quantum Science and Technology, 10(4), 045048.

[33] T. Koike-Akino and Y. Wang, "quEEGNet: Quantum AI for Biosignal Processing," *2022 IEEE-EMBS International Conference on Biomedical and Health Informatics (BHI)*, Ioannina, Greece, 2022, pp. 01-04, doi: 10.1109/BHI56158.2022.9926814.

[34] Nath, R. K., Thapliyal, H., & Humble, T. S. (2021, July). Quantum annealing for automated feature selection in stress detection. In 2021 IEEE Computer Society Annual Symposium on VLSI (ISVLSI) (pp. 453-457).

[35] Torabi, Y., Shirani, S., & Reilly, J. P. (2024). Manikin-Recorded Cardiopulmonary Sounds Dataset Using Digital Stethoscope. arXiv preprint arXiv:2410.03280.

[36] Y. Torabi et al., "Exploring Sensing Devices for Heart and Lung Sound Monitoring," arXiv preprint, doi: 2406.12432, 2024.